\documentclass{article}

\PassOptionsToPackage{numbers, compress}{natbib}

\usepackage[preprint]{neurips_2024}

\usepackage[utf8]{inputenc} 
\usepackage[T1]{fontenc}    
\usepackage{hyperref}       
\usepackage{url}            
\usepackage{booktabs}       
\usepackage{amsfonts}       
\usepackage{nicefrac}       
\usepackage{microtype}      
\usepackage{xcolor}         
\usepackage[ruled,algo2e,noend]{algorithm2e}
\usepackage{algorithm}
\usepackage[english]{babel}
\usepackage{subfigure}
\usepackage{multirow}
\usepackage{wrapfig}

\usepackage{amsmath}
\usepackage{graphicx}
\DeclareMathOperator*{\argmax}{arg\,max}
\DeclareMathOperator*{\argmin}{arg\,min}

\title{Gradient Inversion of Federated Diffusion Models}

\author{%
  Jiyue Huang, Chi Hong \\
  Delft University of Technology \\
  Delft, Netherlands \\
  \texttt{j.huang-4@tudelft.nl} \\
  \texttt{c.hong@tudelft.nl} \\
  \And
  Lydia Y. Chen \\
  University of Neuchatel\\ 
  Neuchatel, Switzerland \\
  \texttt{lydiaychen@ieee.org} \\
  \And  
  Stefanie Roos \\
  RPTU Kaiserslautern \\
  Kaiserslautern, Germany \\
  \texttt{stefanie.roos@cs.rptu.de} \\
}

\newcommand{\alg}{GIDM\xspace}
\newcommand{\algp}{GIDM+\xspace}

\begin{document}

\maketitle

\newcommand{\stef}[1]{{\color{blue}{Stef: #1}}}
\newcommand{\lc}[1]{{\color{olive}{Lc: #1}}}
\newcommand{\gill}[1]{{\color{violet}{Gill: #1}}}

\begin{abstract}
Diffusion models are becoming defector generative models, which generate exceptionally high-resolution image data. Training effective diffusion models require massive real data, which is privately owned by distributed parties. 
Each data party can collaboratively train diffusion models in a federated learning manner by sharing gradients instead of the raw data. In this paper, we study the privacy leakage risk of 
gradient inversion attacks. First, we design a two-phase fusion optimization, \alg, to leverage the well-trained generative model itself as prior knowledge to constrain the inversion search (latent) space, followed by pixel-wise fine-tuning. \alg is shown to be able to reconstruct images almost identical to the original ones.
Considering a more privacy-preserving training scenario, we then argue that locally initialized private training noise $\boldsymbol{\epsilon}$ and sampling step $t$ may raise additional challenges for the inversion attack.
To solve this, we propose a triple-optimization \algp that coordinates the optimization of the unknown data, $\boldsymbol{\epsilon}$ and $t$.
Our extensive evaluation results demonstrate the vulnerability of sharing gradient for data protection of diffusion models, even high-resolution images can be reconstructed with high quality. 

\end{abstract}

\section{Introduction}


The emergence of likelihood-based diffusion models empowers probabilistic models to generate high-quality data, especially for stable high-resolution image and video data~\cite{ddim:conf/iclr/SongME21,ddpm:conf/nips/HoJA20,stablediffusion:conf/cvpr/RombachBLEO22}. 
\begin{wrapfigure}{l}{0.45\textwidth}
\vspace{-1.5em}
  \begin{center}
    \includegraphics[width=0.45\textwidth]{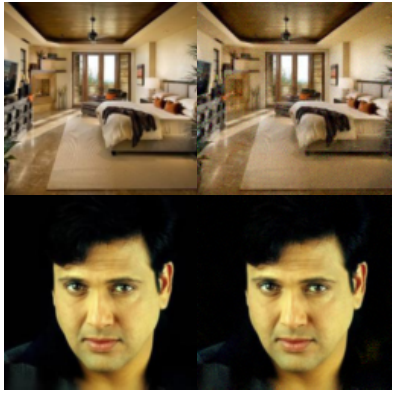}
  \end{center}
  \vspace{-1em}
  \caption{Original training image (left) v.s. Recover images (right) by gradient inversion of diffusion models.}
    \vspace{-2em}
  \label{fig:example_recover}
\end{wrapfigure}
Diffusion models are trained by finding the reverse Markov transitions that maximize the likelihood of the training data. 
In practice, the training is done by gradually adding noise to and denoising the images for multiple steps. 

However, training high-quality diffusion models usually requires a large number of data. Practically, such data may be owned by different private parties. Following data privacy regularization such as HIPAA and GDPR, data owners may keep their data private rather than share it with a central server for collection.
In order to prevent direct local private data collecting but share the knowledge, we are interested in whether the training of diffusion models can be designed by collaborative training without sharing raw data.
Thanks to the design of DDPM at each round, which independently samples a random Gaussian noise $\boldsymbol{\epsilon}$ and a sampling step $t$ to train by approximate noise, we argue that this standard diffusion training paradigm can be translated into the federated learning framework. Specifically, at each global round of training, multiple distributed data owners (clients) train the diffusion sub-models based on their local data. Then, the central server, which connects to every client, aggregates their gradients before sending the gradient back until convergence.



Such a federated training method successfully avoids direct data sharing but opens a door for the server to obtain the intermediate training gradients and the diffusion model in the end. While studies on federated training classifiers demonstrate that the server is able to invert the client's raw data from their intermediate gradients, it is unknown how gradient information also leaks information of the training data in federated diffusion training. Moreover, the well-trained also leak privacy information~\cite{membership:conf/icml/DuanK0SX23,membership:conf/isw/HuP23,memorizing:conf/cvpr/SomepalliSGGG23,memorizing:conf/uss/CarliniHNJSTBIW23}. 
For example, relying on information estimation of the trained model, e.g., error comparison of the forward process posterior
estimation~\cite{membership:conf/icml/DuanK0SX23}, adversaries are able to launch membership inference attacks~\cite{membership:conf/isw/HuP23,membership:conf/icml/DuanK0SX23,membership:journals/corr/abs-2210-00968}. Further, recent studies reveal that trained diffusion models are able to re-create samples that resemble the training data style~\cite{memorizing:conf/cvpr/SomepalliSGGG23}.





In this paper, we systematically study the data reconstruction risk when training federated diffusion models. We first define the federated training method for diffusion models.
Applying the existing inversion attacks for federated classifiers training on diffusion models falls short due to a wide search space of inversion. As the adversarial naturally owns the trained diffusion model and the prior studies~\cite{gias:conf/nips/JeonKLOO21,gifd:conf/iccv/FangCWWX23} show the advantage of including trained generative models, e.g., Generative Adversarial Networks, to improve the reconstruction quality by constraining the searching space of dummy data, we thus advocate to leverage such prior knowledge. Accordingly, we design a fusion optimization, \alg, that includes two training phases. The generating phase maps the dummy data training into a narrow latent space to optimize in-distribution images by adding the diffusion model as prior knowledge. Then the fine-tuning phase further optimizes the similarity between the dummy and real gradients to update the dummy data generated by the generating phase. Our proposed \alg is able to recover high-resolution images of $128 \times 128$ almost identical to the origin, as shown by Fig.~\ref{fig:example_recover}, where uncertainty by searching space increases exponentially with increased pixel size.

Furthermore, we consider a more challenging, privacy-preserving training scenario in which the data owner samples an arbitrary timestep $t$ and initializes $\epsilon$ privately and locally, i.e., the honest-but-curious server is unaware of.
To solve this, we then propose the novel triple optimization \algp for dummy data, $\epsilon$ and $t$. Specifically, the triple optimization includes three independent optimizers for the large-tensor dummy data, $\boldsymbol{\epsilon}$ following Gaussian distribution for initialization, and the integral $t$ with long range to refine the joint training. By coordinating the optimization of them with updating intervals, we are able to recover images without confidently knowing $\epsilon$ and $t$.

We summarize our main contribution as follows:

\begin{itemize}
    \item We are the first study on the gradient inversion attack for private training data of federated diffusion models. 
    \item We design a two-phase fusion optimization \alg that effectively inverts the client gradients into images by leveraging the prior knowledge of trained diffusion models. \alg is able to re-create almost identical images of the training data for size $128 \times 128$.
    \item To further tackle the more privacy-preserving scenario when $\epsilon$ and $t$ are private to clients, we design a triple optimization \algp that simultaneously inverts clients' images, and private training parameters.  
\end{itemize}

\section{Related Work}
\textit{Diffusion Models and Privacy.}
Diffusion models employ a two-step process: initially deconstructing the training data structure step by step in a forward manner, then mastering the reconstruction of the structure from noise in a reverse process. The Denoising Diffusion Probabilistic Model (DDPM)~\cite{ddpm:conf/nips/HoJA20} introduced the stable and efficient implementation for high-quality image synthesis. It is achieved via a forward process without learnable parameters while employing simplified Gaussian noise in the reverse phase. Further variants of diffusion models such as DDIM~\cite{ddim:conf/iclr/SongME21}, Stable Diffusion~\cite{stablediffusion:conf/cvpr/RombachBLEO22}, and Imagen~\cite{imagen:conf/nips/SahariaCSLWDGLA22} thus improve the sampling efficiency or involve deep language understanding for text-to-image generation. However, The well-trained diffusion model network have been shown to be vulnerable to privacy attacks, i.e., training data
information leakage. Recent studies on privacy concerns of diffusion models mainly focus on membership inference~\cite{membership:conf/icml/DuanK0SX23,membership:conf/isw/HuP23,membership:journals/corr/abs-2210-00968} attacks or training data memorizing attacks~\cite{memorizing:conf/cvpr/SomepalliSGGG23,memorizing:conf/uss/CarliniHNJSTBIW23} from the trained model. None of the studies has addressed the privacy leakage impact from the gradients of diffusion models. 

\textit{Gradient Inversion.}
As the first practical gradient inversion attack for classifiers, Deep leakage from Gradients (DLG) reconstructs data and label simultaneously by directly approximating gradients from the dummy data input~\cite{dlg:conf/nips/ZhuLH19}. Two main aspects that DLG suffers from are reconstructed data quality and the inability to deal with large training batches. To strengthen DLG, one line of work improves DLG by developing different optimizer~\cite{agic:conf/srds/XuHHCD22,invG:conf/nips/GeipingBD020}, distance measure metrics~\cite{invG:conf/nips/GeipingBD020}, or integrating direct features~\cite{idlg:journals/corr/abs-2001-02610} by the network, e.g., they first infer labels before reconstructing. The other line of work focuses on bringing external knowledge for inversion. Such knowledge includes knowing prior data distributions for more accurate embedding,
adding a batch normalization
regularizer to manage larger reconstructing batches~\cite{gradinversion:conf/cvpr/YinMVAKM21}, applying pre-trained generator model to ensure high-quality output of reconstructed data~\cite{gias:conf/nips/JeonKLOO21,gifd:conf/iccv/FangCWWX23,GGL:conf/cvpr/Li0L022} or even utilizing auxiliary dataset. Besides classification models, although generative models assist the inversion in general, inversion also appears in Generative Adversarial Networks (GANs), which aim to invert a given image back into the latent space of a pre-trained GANs model~\cite{GANinv:journals/pami/XiaZYXZY23}. Currently, there is no inversion attack studied for diffusion models, namely reconstructing the input data from the intermediate gradients of the federated diffusion model training.

\section{Methodology}

In this section, we first define a federated diffusion training method that enables distributed learning without sharing data partition 
We then define and propose a gradient inversion attack for federated diffusion models, which composes two core optimization algorithms. 
The first algorithm is a constrained inversion attack, focusing on a scenario where all training hyperparameters are identical across clients. It specifically leverages prior knowledge of pretrained diffusion models to narrow down the inversion searching space.  
The second algorithm considers a privacy-preserving federated method that two key hyperparameters of diffusion model training, $\{\boldsymbol{\epsilon}, t\}$, are sampled privately by local clients. We thus propose a triple optimization algorithm, which optimizes and reconstrcuts client data, and $\{\boldsymbol{\epsilon}, t\}$ simultaneously. 

\subsection{Federated diffusion models}
\label{subsec:federated_diffusion}
\begin{algorithm}[t]
\SetAlgoLined
  \caption{Federated Diffusion Model Training} 
  \label{alg:code}
    \textbf{Input:} 
    The number of clients $K$, number of global training round $R$, local datasets $X_k, k \in [1, K]$, diffusion steps $T$, global learning rate $\eta$.

    Initialize model $\theta$\\
    \For{$r = 1, 2, ..., R$}
    {
        \For{$k = 1, 2, ..., K$}
        {
         {$t \sim \text{Uniform}(\{1,...,T\})$ 
         \tcp*[f]{Excuted by the server}\\
         }
        {$\boldsymbol{\epsilon} \sim \mathcal{N}(\boldsymbol{0}, \boldsymbol{I})$ 
         \tcp*[f]{Excuted by the server}\\
         }
        {$\boldsymbol{x}_0 \sim X_k$
        \tcp*[f]{Excuted by the client}\\
        Calculate $g_k = \nabla_{\theta_k}\left\|\boldsymbol{\epsilon}_k-\boldsymbol{\epsilon}_{\theta_k}\left(\sqrt{\bar{\alpha}_t} \mathbf{x}_0+\sqrt{1-\bar{\alpha}_t} \boldsymbol{\epsilon}_k,t\right)\right\|^2$
        \tcp*[f]{Excuted by the client}\\}
        }
        $g = \sum_{k=1}^K{g_k}$
        \tcp*[f]{Excuted by the server}\\
        Update $\theta = \theta - \eta g$
        \tcp*[f]{Excuted by the server}\\
    }
    {Return $\theta$}
    \\
    \textbf{Result:} The trained $\theta^*$
\end{algorithm}

We consider a federated image generative task following the standard DDPM-like diffusion model~\cite{ddpm:conf/nips/HoJA20}, which aims at optimizing weighted variational bound:
\begin{equation}
\label{eq:dm_objective}
    L(\theta)=\mathbb{E}_{t, \mathbf{x}_0, \boldsymbol{\epsilon}}\left[\left\|\boldsymbol{\epsilon}-\boldsymbol{\epsilon}_\theta\left(\sqrt{\bar{\alpha}_t} \mathbf{x}_0+\sqrt{1-\bar{\alpha}_t} \boldsymbol{\epsilon}, t\right)\right\|^2\right],
\end{equation}

where $\boldsymbol{x}_0 \in \mathbb{R}^{m \times B}$ is the image training samples (batch size $B$) of $m=(\text{width}) \times (\text{height}) \times (\text{color})$, $L(\cdot)$ is the point-wise loss function, $\theta$ denotes the diffusion model network parameters and $\bar{\alpha}_t$ is hyper-parameter controlling the forward noising process.
Note that the sampling step $t$ is uniform between 1 and $T$ and we follow the definition~\cite{ddpm:conf/nips/HoJA20} that $\boldsymbol{\epsilon}_\theta$ is a function approximator intended to predict the added Gaussian noise $\boldsymbol{\epsilon}$ to the image of step $t$, $\boldsymbol{x}_t$.

There are $K$ clients serving as data owners and are responsible for diffusion model training. The $k^{th}$ federated learning client owns the local real dataset $X_k, k \in \{1, 2, ..., K\}$, which is not shared with others. Each client reports the gradient of $\nabla_\theta\left\|\boldsymbol{\epsilon}_k-\boldsymbol{\epsilon}_{\theta_k}\left(\sqrt{\bar{\alpha}_t} \mathbf{x}_0+\sqrt{1-\bar{\alpha}_t} \boldsymbol{\epsilon}_k, t\right)\right\|^2$ for the sampled data $\boldsymbol{x}_0$ sampled locally. We introduce a $R$ global training round of training. The single server $\mathcal{S}$, which does not own any data itself, aggregates and distributes the gradient on each global training round. 

One canonical difference between federated diffusion and other models, e.g., classification tasks, lies in which party is supposed to sample $\epsilon$ and $t$ for a given $\boldsymbol{x}_0$. Specifically, it can be either the client that samples $\{\epsilon, t\}$ locally and privately for training each batch, or the server, which samples $\{\epsilon, t\}$ and sends them to each client per round per batch together with the aggregated model. Both settings can be implemented as equivalent optimization problems of centralized diffusion models. We consider two scenarios regarding whether the information of $\{\epsilon, t\}$ is kept as clients' private information. In the first scenario, $\{\epsilon, t\}$ are sampled by the honest-and-curious server and send them to each client per batch training. Given the global model learning rate as $\eta$, the training of the federated diffusion model is equivalent to the centralized DDPM and is illustrated in Alg.~\ref{alg:code}. 
Then, we extend to the scenario where $\{\epsilon, t\}$ are kept privately.

\subsection{\alg: Constrained gradient inversion attack by diffusion prior}

To study the privacy leakage risk of the federated diffusion based on gradients, we propose a gradient inversion attack. The honest-but-curious server reconstructs the victim client's data using the gradients submitted by the client and known values of $\epsilon$ and $t$. Our detailed description of the threat model is included in Appendix.~\ref{app:threat}. We model the inversion process as an optimization problem that iteratively optimizes the dummy data by minimizing the distance between known and approximate gradients, as proposed by the inversion attacks for federated classifier training~\cite{dlg:conf/nips/ZhuLH19}.
Following Alg.~\ref{alg:code}, when client $k$ computes the gradient by a batch of data $\boldsymbol{x}_0 = \{x_1, x_2, ..., x_B\}$ with batch size $B$, the gradient is formulated as $g_k = \frac{1}{B}\sum_{i=1}^B \nabla_{\theta_k}\left\|\boldsymbol{\epsilon}_k-\boldsymbol{\epsilon}_{\theta_k}\left(\sqrt{\bar{\alpha}_t} x_i +\sqrt{1-\bar{\alpha}_t} \boldsymbol{\epsilon}_k,t\right)\right\|^2$. 
Note that since a gradient inversion attack can be launched at any specific global training round for any client, we ignore the $r$ round and $k$ client indexes in the following section. 
Assuming that $\theta$ is second-order differentiable, we suppose that dummy data $\hat{\boldsymbol{x}}_0$ is the replica of $\boldsymbol{x}_0$ if $\hat{g} \sim g$, where $\hat{g}$ is the dummy gradient calculated based on dummy $\hat{\boldsymbol{x}}_0$.
Thus, our gradient inversion objective turns to:
\begin{equation}
\label{eq:dm_inversion}
   \min_{\hat{x}_1, ..., \hat{x}_B}{D}(\frac{1}{B} \sum_{i=1}^B \nabla_{\theta}\left\|\boldsymbol{\epsilon}-\boldsymbol{\epsilon}_{\theta}\left(\sqrt{\bar{\alpha}_t} \hat{x}_i +\sqrt{1-\bar{\alpha}_t} \boldsymbol{\epsilon},t\right)\right\|^2, g) 
\end{equation}

where ${D}(\cdot)$ is the distance measure metric for two gradients to show similarity. We define $D(\cdot)$ by L2-Norm~\cite{dlg:conf/nips/ZhuLH19} distance.
\begin{figure}
    \centering
    \includegraphics[width=1\textwidth]{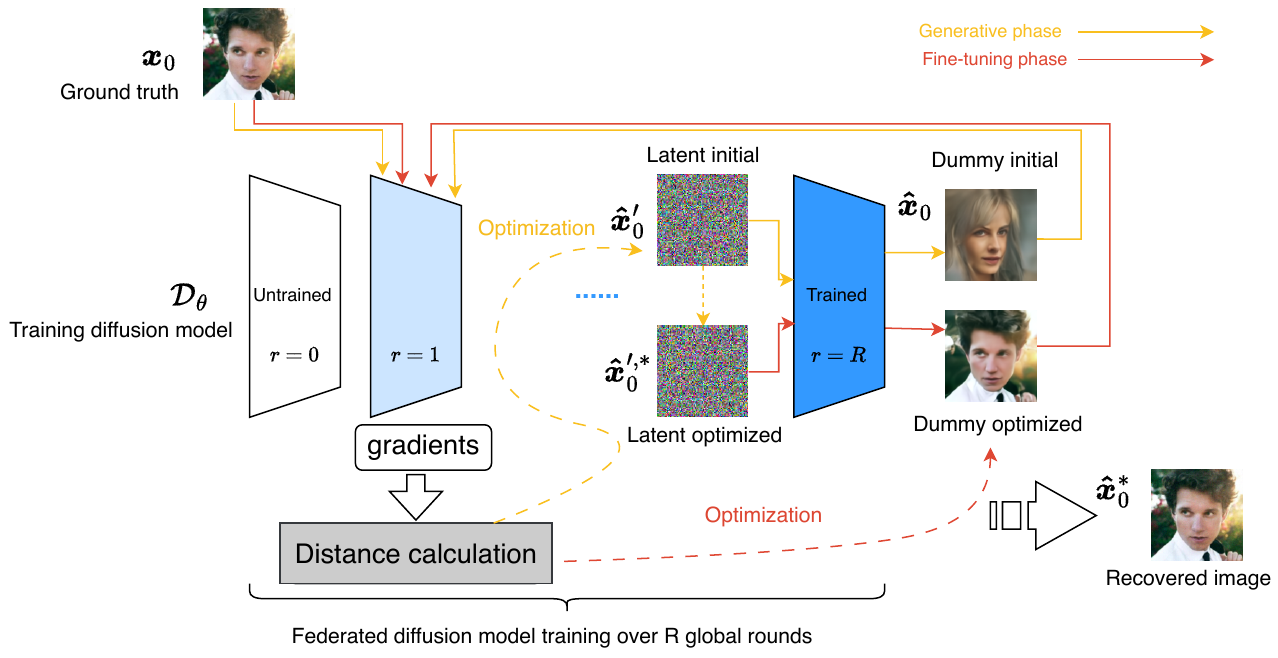}
    \caption{Gradient inversion on diffusion models with constrained space by the trained diffusion models. We utilize one $128 \times 128$ figure from \textit{CelebA} dataset as an example to show every intermediate result above. The fusion optimization includes two phases: the generative phase (8000 iterations on this example) and the fine-tuning phase (1500 iterations), to reconstruct with efficiency and effectiveness. The output of the generative phase differs from the original data by hair, face, and background.}
    \label{fig:fusion_optimization}
\end{figure}

We first conduct an exploratory experiment on whether the direct translation from the inversion method of the classifier can launch a successful gradient inversion attack on diffusion models. The result presented in Sec.~\ref{subsec:main_result} demonstrates their inability to recover training samples. The reason behind this lies in the wide search space of the optimization procedure. Also, the difficulty of inversion in terms of randomness increases greatly with high-resolution data.

To improve the optimization for high-resolution data, we propose a constrained fusion optimization to leverage prior knowledge of the trained diffusion model,
$\mathcal{D}_\theta^*: \mathbb{R}^m \rightarrow \mathbb{R}^m$, where $\theta^*$ is the diffusion model parameter. 
Our assumption is that the server naturally possesses the trained diffusion model. Integrating the knowledge of such a trained diffusion model brings the advantage of constraining the searching space in optimizing the dummy data. Our proposed fusion optimization consists of two phases: the generative phase and the fine-tuning phase.

\textbf{Generative phase. }We aim to generate images that resemble the training dataset as the starting point of dummy data optimization by gradient approximation. It is achieved by mapping the original search into a narrow latent space with constraints (or prior knowledge). Let the latent code $\hat{\boldsymbol{x}}^{\prime}_0$ be the same dimension of the dummy data $\hat{\boldsymbol{x}}_0$ from Eq.~\ref{eq:dm_inversion} and $\hat{\boldsymbol{x}}_0 = \mathcal{D}_\theta(\hat{\boldsymbol{x}}^{\prime}_0, t)$, we execute to optimize $\hat{\boldsymbol{x}}^{\prime}_0$ multiple iterations so that the gradient computed on $\hat{\boldsymbol{x}}_0$ has a small distance with the real gradient. By this means, instead of directly updating the dummy data, we perform latent space search over $\boldsymbol{\hat{x}}^{\prime}_0$, which is the input of $\mathcal{D}_\theta$, outputting $\boldsymbol{\hat{x}}_0$. That is: 
$\boldsymbol{\hat{x}}_0^{\prime,*} = \mathcal{D}_\theta(\left.\argmin_{\hat{\boldsymbol{x}}^{\prime}_0} D (\left.\nabla_\theta\left\|\boldsymbol{\epsilon}-\boldsymbol{\epsilon}_\theta\left(\sqrt{\bar{\alpha}_t}\mathcal{D}_\theta(\hat{\boldsymbol{x}}^{\prime}_0, t) +\sqrt{1-\bar{\alpha}_t} \boldsymbol{\epsilon}, t\right)\right\|^2, g\right), t\right)$.

\textbf{Fine-tuning phase. } The generative phase is able to generate high-quality data, whereas indirectly optimizing $\hat{\boldsymbol{x}}_0$ does not guarantee pixel-wise similarity. On the other hand, each iteration of optimizing the latent code $\hat{\boldsymbol{x}}^{\prime}_0$ requires $T$ sampling steps of the trained diffusion model, e.g., $T=1000$ in DDPM, which is computationally intensive. Therefore, our fine-tuning phase executes direct optimizing of $\hat{\boldsymbol{x}}_0$ following Eq.~\ref{eq:dm_inversion} based on the output of the generative phase $\boldsymbol{\hat{x}}_0^{\prime,*}$ to increase the pixel-wise similarity.

Our method is illustrated in Fig.~\ref{fig:fusion_optimization}. 
The output of the generative phase, i.e., the intermediate dummy data, generated by the optimized latent code $\boldsymbol{\hat{x}}_0^{\prime,*}$, is then optimized by the fine-tuning phase. From Fig.~\ref{fig:fusion_optimization} we see that the generative phase is already able to reconstruct a high-quality image of resembled style while the pixel-wise similarity with the original data is low. 
By integrating the generative phase and the fine-tuning phase, we recover high-resolution data efficiently and effectively. 


\subsection{\algp: Triple-optimization for private $\epsilon$ and $t$}
\label{subsec:triple_optimization}

We now consider the federated diffusion training that both Gaussian noise and sampling step are private information that are only known by the clients.
To implement the inversion without knowing $\{ \boldsymbol{\epsilon}, t\}$,  we need to optimize them three simultaneously while considering their design principles and differences. 
In this part, we design a triple-optimization method to refine the coordination of the three independent optimizations, namely, $\boldsymbol{x}_0$, $\epsilon$, and $t$. Note that all of their optimizations are based on backpropagating for approximating the gradients. 



\textbf{Optimizing $\hat{\boldsymbol{\epsilon}}$.}  
In DDPM, we compute the distribution of the noisy sample after $t$ iterations of the forward process in closed-form by $q\left(\mathbf{x}_t \mid \mathbf{x}_0\right)=\mathcal{N}\left(\mathbf{x}_t; \sqrt{\bar{\alpha}_t} \mathbf{x}_0,\left(1-\bar{\alpha}_t\right) \mathbf{I}\right)$. The number of iterations in the forward process is set to a large $T$ 
, and the variance levels $\beta_t \in (0, 1), \bar{\alpha}_t=\prod_{i=1}^t\left(1-\beta_t\right)$ increase linearly (ranging from $10^{-4}$ to $0.02$), which means that $\bar{\alpha}_t \rightarrow 0$, and the latent distribution is a Gaussian distribution
$\epsilon$ sampled locally by the client should follow Gaussian distribution $\boldsymbol{\epsilon} \sim \mathcal{N}(\boldsymbol{0}, \boldsymbol{I})$. When implementing stochastic gradient descent, it requires a small learning rate $\eta_{\epsilon}$ to update $\boldsymbol{\hat{\epsilon}}$. 
Following Eq.~\ref{eq:dm_objective}, training diffusion models is to predict noises
added in the forward process and performs well with a stable noise.
Thus, we utilize an interval updating strategy that after $S$ iterations of updating $\boldsymbol{x}_0$ and $t$, we perform $S$ iterations of $\boldsymbol{\hat{\epsilon}}$, $\boldsymbol{x}_0$ and $t$ simultaneously:
    $\boldsymbol{\hat{\epsilon}}^{i+1} =  \boldsymbol{\hat{\epsilon}}^{i} - \eta_{\epsilon} \frac{\partial D(\hat{g}, g)}{\partial \boldsymbol{\hat{\epsilon}}^{i}}, \, s.t. i \% 2S >S. $

\textbf{Optimizing $t$.} Different from $\boldsymbol{\epsilon}$ and $\boldsymbol{x}_0$, which are floating tensors, $t$ is an integer ranging from $[1, T]$. To find the optimal discrete $t$ by stochastic gradient descent, we define and initialize a $1 \times T$ auxiliary vector following uniform distribution: $\boldsymbol{\hat{t}} \sim \text{Uniform} (\boldsymbol{0}, \boldsymbol{J})$. Each iteration of optimization by approximating dummy gradient to real gradient, $\boldsymbol{\hat{t}}$ is updated by learning rate $\eta_t$. The inferred $\hat{t}$ is designed as the index of the maximum element in the vector after softmax transition: $\hat{t} = \argmax (softmax (\boldsymbol{\hat{t}}))$. $\hat{t}$ is updated at each inversion iteration:
$    \hat{t}^{i+1} = \hat{t}^{i} - \eta_{t} \frac{\partial D(\hat{g}, g)}{\partial \hat{t}^{i}}.
$

\textbf{Optimizing $\boldsymbol{\hat{x}}_0$.} Based on the dummy $\boldsymbol{\hat{\epsilon}}$ and $\hat{t}$ as $\boldsymbol{\epsilon}$ and $t$ Eq.~\ref{eq:dm_inversion}, performing the optimization of $\hat{\boldsymbol{x}}_0$ starts with a random initialized image $\hat{\boldsymbol{x}}_0 \sim \text{Uniform}(\boldsymbol{0}, 2^p\boldsymbol{I})$, given that the source data is a $p$-bit binary encoded value. With each iteration, we update $\hat{\boldsymbol{x}}_0$ by back-propagating the distance between $\hat{g}$ and $g$. We set the learning rate of the dummy data as $\eta_x$. $\boldsymbol{x}_0$ is updated at each inversion iteration: $\hat{t}$ is updated at each inversion iteration:
$    \hat{\boldsymbol{x}}_0^{i+1} = \hat{\boldsymbol{x}}_0^{i} - \eta_{t} \frac{\partial D(\hat{g}, g)}{\partial \hat{\boldsymbol{x}}_0^{i}}.
$
For all three optimizations, we use Adam as the optimizer.

In summary, the triple-optimization gradient inversion coordinates three independent optimization process to perform data reconstruction on more privacy-preserving federated diffusion models. 








\section{Experiment}
\label{sec:experiment}

\textbf{Setups.} Unless stated otherwise, our experiments are implemented on two datasets for the scenario when the adversarial server knows $\{ \boldsymbol{\epsilon}, t\}$:  \textit{Celeb-A}~\cite{celebA:conf/iccv/LiuLWT15}and \textit{LSUN-Bedroom}~\cite{lsun:journals/corr/YuZSSX15}, both  with image size resized to $128 \times 128$. Both datasets are trained using the standard DDPM model~\cite{ddpm:conf/nips/HoJA20} as by Alg.~\ref{alg:code} with 50 rounds. For the challenging case when $\{ \boldsymbol{\epsilon}, t\}$ are private to clients, we use \textit{Cifar-100}~\cite{cifar:krizhevsky2009learning} dataset with image size $32 \times 32$.
The optimization interval of \algp for $\boldsymbol{\epsilon}$ optimizer is set to be 50. For measuring the image similarity between the recovered data and the original, we use four evaluation metrics: MSE~\cite{mse:bickel2015mathematical}, PSNR~\cite{psnr:conf/icpr/HoreZ10}, SSIM~\cite{ssim:journals/tip/WangBSS04}, and LPIPS~\cite{lpips:conf/cvpr/ZhangIESW18}. The detailed setups and datasets are presented in Appendix.~\ref{app:setup} and the description of our evaluation metrics in Appendix.~\ref{app:metric}.

\textbf{Baselines.} Since we are the first to study the gradient inversion attack on diffusion models, there is no direct baseline to compare with. Most inversion methods are designed for classifiers. Thus, we compare our data reconstructing results with adapted versions of the attacks that are compatible with diffusion models with single batch size: DLG-dm~\cite{dlg:conf/nips/ZhuLH19} and InvG-dm~\cite{invG:conf/nips/GeipingBD020}  with Adam optimizer. The details on how to translate DLG and InvG for classifiers to DLG-dm and InvG-dm for diffusion models are explained in Appendix.~\ref{app:setup}.

\subsection{Final reconstructed images}
\label{subsec:main_result}

\begin{figure}
    \centering
    \includegraphics[width=1.0\textwidth]{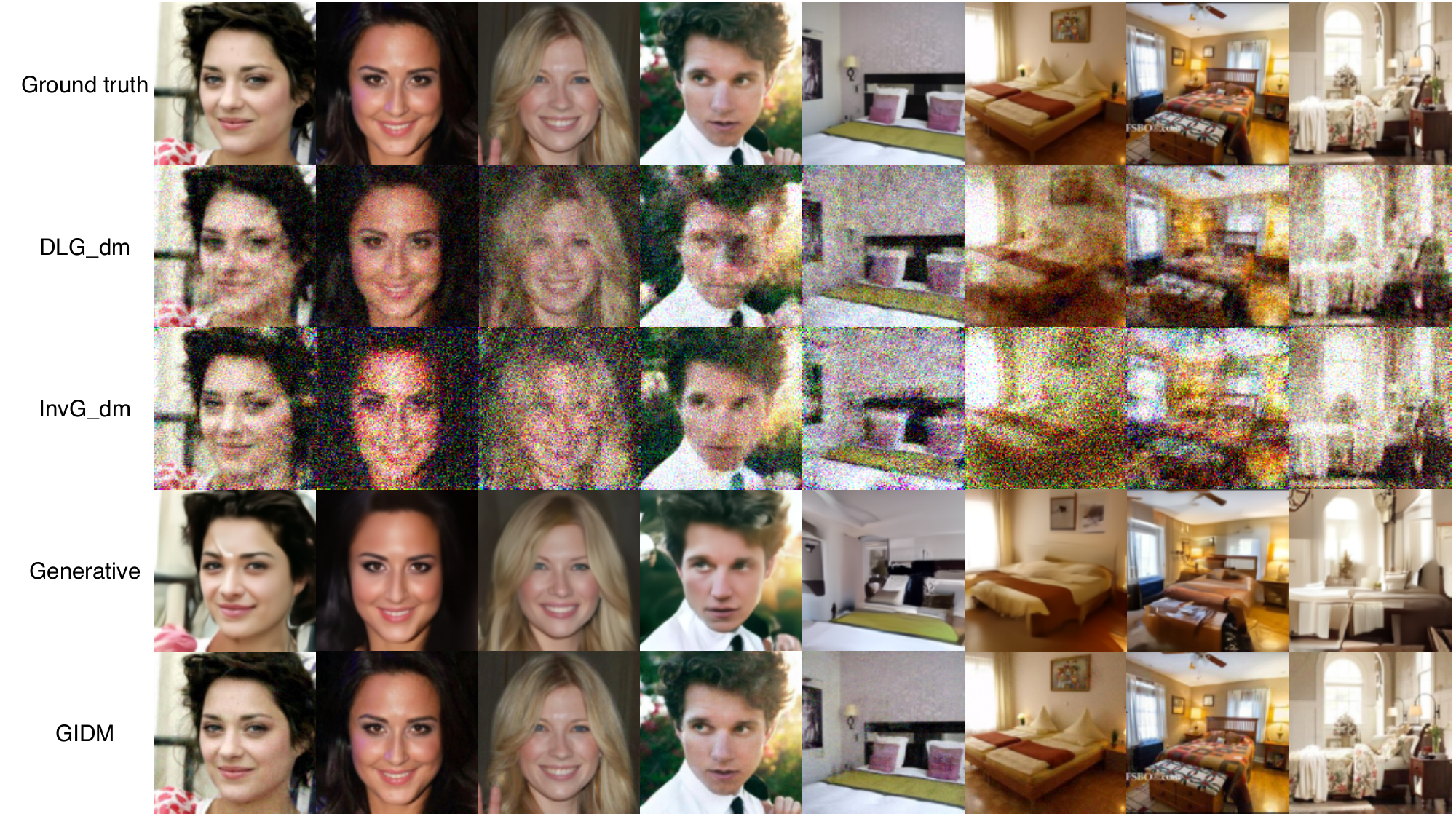}
    \caption{Visualization of recovered images comparing with baselines on \textit{CelebA} and \textit{LSUN-Bedroom} datasets. Images reconstructed with server initializing $\{\boldsymbol{\epsilon}, t\}$ are $128 \times 128$ while with the private client initialization, $32 \times 32$ images are successfully recovered. ``Generative'' in the figure is the output of our generative phase before the fine-tuning phase.}
    \label{fig:visualization_main}
\end{figure}

To demonstrate our inversion effectiveness, we report the final MSE, LPIPS, PSNR, SSIM results in Tab.~\ref{tab:inversion_main} (when the server initializes $\{\boldsymbol{\epsilon}, t\}$) and Tab.~\ref{tab:inversion_unknown} (when the client keep them private locally). We also visualize the final reconstructed image examples in Fig.~\ref{fig:visualization_main} for \textit{CelebA} and \textit{LSUN-bedroom} reconstruction. $\downarrow$ stands for the lower the better while $\uparrow$ stands for the higher the better quality of reconstruction based on the metric definitions. We discuss both the case when $\{\boldsymbol{\epsilon}, t\}$ is initialized by the server and when it is privately initialized by the client. When $\{\boldsymbol{\epsilon}, t\}$ is sampled by the client, we assume that the baseline methods randomly initialize them: $\hat{\boldsymbol{\epsilon}} \sim \mathcal{N}(\boldsymbol{0},\boldsymbol{I}) $ and $\hat{t} \sim \text{Uniform}(\{1,...,T\})$. ``Generative'' in Tab.~\ref{tab:inversion_main} refers to the output of the generative phase before the fine-tuning phase, reported as an ablation study of our \alg.

We start with the case when the server initializes $\{\boldsymbol{\epsilon}, t\}$. From the Tab.~\ref{tab:inversion_main}, it is evident that \alg outperforms baseline methods significantly over four different evaluation metrics, demonstrating superior reconstructing results. DLG-dm and InvG-dm have different loss functions in terms of measuring the similarity between the dummy gradients and the real. 
Another point worth mentioning is that the output of our generative phase does not always recover better images than DLG-dm and InvG-dm. It can be explained by the fact that our four evaluation metrics compute the quality of similarity based on pixels. ``Generative'' optimizes the latent space to conduct indirect inversion. Thus, the high-quality output (semantically similar and clear) from the diffusion model may result in lower pixel-wise similarity than baselines, which is even blurred by Fig.~\ref{fig:visualization_main}.

\begin{table}[th!]
\centering
\caption{Results for inversion of diffusion models on \textit{CelebA} and \textit{LSUN-Bedroom} datasets with $\{\boldsymbol{\epsilon}, t\}$ initialized by the server. We compare our \alg with the baseline methods adapted to diffusion inversion for two initialization scenarios.}
\begin{tabular}{c|c|r|r|r|r} 
\toprule
Dataset & Method& \textbf{MSE} $\downarrow$ & \textbf{LPIPS} $\downarrow$ & \textbf{PSNR} $\uparrow$ & \textbf{SSIM} $\uparrow$ \\
\midrule
\multirow{4}{*}{\textit{CelebA}} &DLG-dm& 1.59e-2 & 0.67 & 66.12 & 0.39 \\
 &InvG-dm& 0.04 & 0.75 & 62.43 & 0.30 \\
 &Generative (ours) & 5.99e-3 & 0.11 & 70.36 & 0.71 \\
 &\textbf{\alg} (ours) & \textbf{7.31e-4} &\textbf{0.04} &\textbf{79.49} & \textbf{0.91} \\

\midrule
\multirow{4}{*}{\textit{LSUN-Bedroom}} &DLG-dm& 0.02 & 0.64 & 65.37 & 0.37 \\
 &InvG-dm & 3.71e-2 & 0.73 & 62.43 & 0.26 \\
 &Generative (ours) & 0.01 & 0.26 & 66.74 & 0.54 \\
 &\textbf{\alg} (ours)& \textbf{7.13e-5} & \textbf{3.44e-3} & \textbf{89.60} & \textbf{0.99} \\
\bottomrule
\end{tabular}

\label{tab:inversion_main}
\end{table}

Fig.~\ref{fig:visualization_main} visualizes and compares the recovered images on both datasets of size $128 \times 128$. 
\alg successfully reconstructs high-resolution images from the gradients that are nearly indistinguishable from the ground truth perceptually. 
As a comparison, DLG-dm and InvG-dm, which are designed for classifiers, fail to recover resembled images. Specifically, they are only able to create images of similar color palettes as the original data without recreating the original object, let alone high-resolution details. This is within our expectations since they conduct pixel-wise optimization. Such a way suffers from exponentially increased difficulty of searching given a larger image size, even when $\{\boldsymbol{\epsilon}, t\}$ are known by the adversarial server.

When it comes to the difference between ``Generative'' and \alg, we observe that our generative phase with the assistance of diffusion prior knowledge can already recover high-quality images. Generally, the main color, object outline, and positions after the generative phase are almost identical to the ground truth, though there are still minor differences in the details of the images. For example, the face and hair shape, the background texture, or sometimes the makeup color is different from the ground truth on the \textit{CelebA} human-facial dataset. On \textit{LSUN-bedroom}, we can see more distinct differences in objects such as door and lap. Motivated by this, our next step of the fine-tuning phase adjusts the generated images by direct gradient approximation, outputting \alg results of successful reconstruction.

Next, we move to the case when  $\{\boldsymbol{\epsilon}, t\}$ is kept private by the clients. 
As the classifier baselines DLG and InvG do not have the intentional design for $\boldsymbol{\epsilon}$ and $t$, which is specified by diffusion probabilistic models, we use unchanged Gaussian random initialization for their $\boldsymbol{\epsilon}$ and uniformly sampled $t$ from $[1, T]$.
The results on \textit{Cifar-10} datasets are presented in Tab.~\ref{tab:inversion_unknown}. From the results, \algp shows obvious superiority against DLG-dm and InvG-dm, demonstrating the effectiveness of our design of triple-optimization, refining three independent optimizers.
Both baseline methods fail to reconstruct in terms of the similarity measure metrics. 
We may explain this by the process that the training of diffusion models defined by Eq.~\ref{eq:dm_objective} predicts the sampled noise for training each step. However, once the noise is kept confidential, the random initialized noise by DLG-dm and InvG-dm, which is different from the one from training, will lead to unsuccessful inversion due to totally different optimization. For \algp, optimizing three independent optimizers with intervals provides information to each other for coordination. However, \algp also finds it challenging when recover high-resolution images. We will leave this to further work.

\begin{table}[th!]
\centering
\caption{Results for inversion of diffusion models on \textit{Cifar-10} datasets with $\{\boldsymbol{\epsilon}, t\}$ initialized by the clients and is unknown to the adversary. We compare \algp with the baseline methods adapted to diffusion inversion for two initialization scenarios.}
\begin{tabular}{c|r|r|r|r} 
\toprule
 Method& \textbf{MSE} $\downarrow$ & \textbf{LPIPS} $\downarrow$ & \textbf{PSNR} $\uparrow$ & \textbf{SSIM} $\uparrow$ \\
\midrule
DLG-dm& 69.01 & 0.34 & 29.74 & 4.62e-3 \\
InvG-dm& 1.30 & 0.32 & 47.00 & 0.25 \\
\textbf{\algp} (ours) & \textbf{0.36} &\textbf{0.14} &\textbf{52.61} & \textbf{0.54} \\
\bottomrule
\end{tabular}

\label{tab:inversion_unknown}
\end{table}

To summarize, gradient inversion of diffusion models by integrating the trained generator as prior knowledge to optimize the latent space greatly constrained the search space of the high-resolution images, demonstrated by generating high-quality data similar to the ground truth. However, such indirect inversion by optimizing latent code while outputting images from the generative model does not guarantee a small pixel-wise distance. Thus, a followed fine-tuning phase that conducts direct pixel-wise optimization according to approximate gradients further enhances the data reconstruction attack, demonstrated by generated images that are visually nearly identical to the original.

\subsection{Intermediate inversion outputs on \alg}

\begin{figure}[th]
    \centering
    \includegraphics[width=1.0\textwidth]{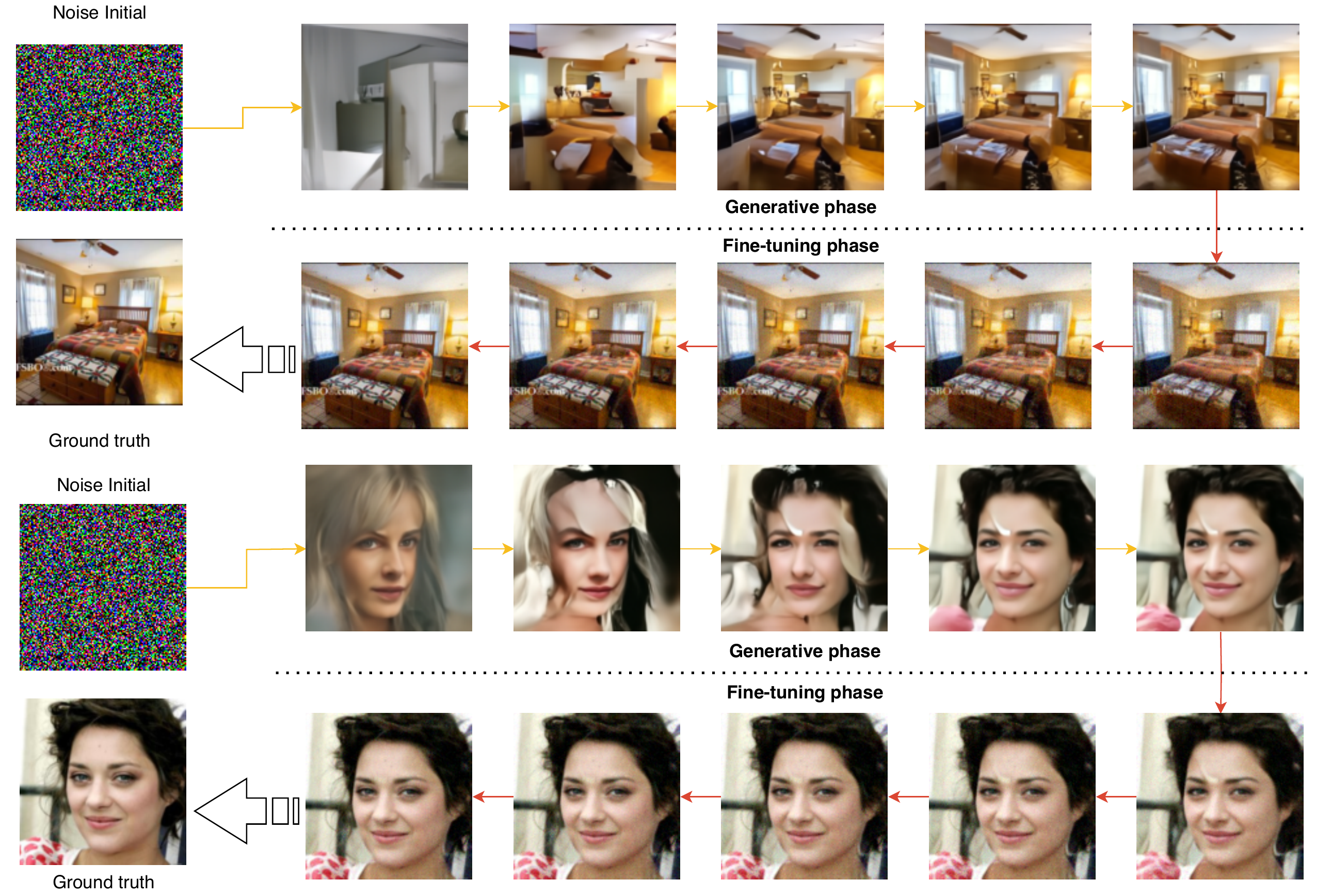}
    \caption{Step-by-step gradient inversion intermediate results visualization. The generative phase is marked by yellow arrows while the fine-tuning phase is red arrows. We initialize the latent code by $128 \times 128$ Gaussian noise. The generative phase is trained by 5000 iterations while the fine-tuning phase by 2000 iterations. We present the evenly divided iterations through the whole process.}
    \label{fig:visual_intermediate}
\end{figure}

Let us zoom into how the image is recovered by our proposed \alg and bring some insights on the vulnerability of federated diffusion models against gradient inversion attacks. We use $128 \times 128$ image example from \textit{LSUN-Bedroom} and \textit{CelebA} dataset, illustrated in Fig.~\ref{fig:visual_intermediate}.

From the figure, for the generative phase, the initialized random Gaussian noise can output a decent image of a bedroom. Although it is exactly different from the ground truth image, it has constrained the search space of dummy optimization by the knowledge of the well-trained diffusion model.
After calculating the dummy gradient based on the previous bedroom image, our back-propagation by approximating the dummy to the real gradient adjusts the Gaussian noise so that it gradually generates an image of a similar color palette and object outlines. One interesting finding is that the color turns similar at a very early iteration while the object outlines converge slowly and gradually, such as the hanging paintings. Upon reaching the 5000-th iteration of the generative phase, the recovered image has been gradually adjusting the details in the image. Yet, we have tried to increase the number of iterations to 15000 and found the final image still cannot fully recover the ground truth
due to this indirect optimization process. The finding motivates us to integrate direct updating. This is where our fine-tuning phase comes into play.

The fine-tuning phase starts with the output of the generative phase and we train 1800 iterations. Following the red arrows, the blanket, and hanging paintings are turning to approximate the original data. The fine-tuning phase does not utilize the diffusion model for optimizing alternative space. Thus, the reconstruction is efficient due to unnecessarily executing diffusion model sampling at each optimization iteration as the generative phase.

\subsection{The impact of $t$ on \algp}

\begin{figure}[th]
    \centering
    \includegraphics[width=1.0\textwidth]{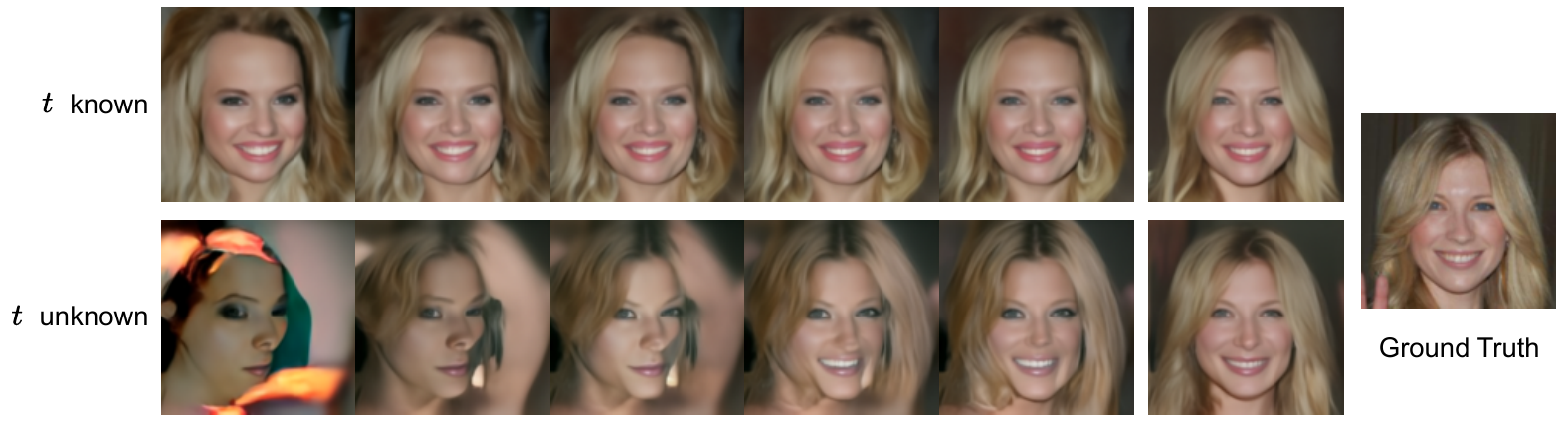}
    \caption{Step-by-step gradient inversion intermediate results visualization, comparing known or unknown $t$. We initialize the latent code by $128 \times 128$ Gaussian noise. We present some of the representative intermediate outputs through the whole process. The total number of inversion iterations is 3000.}
    \label{fig:t_effectiveness}
\end{figure}

Our results in Sec.~\ref{subsec:main_result} consider $\boldsymbol{\epsilon}$ and $t$ initialized together by either the server or the client. Here we provide an ablation study on the impact of knowing $t$. We have evaluated that it is difficult to recover high-resolution images of high quality without knowing $\boldsymbol{\epsilon}$ due to a high order of randomness brought by predicting the high-dimensional noise. We use an image from \textit{CelebA} as an example and show the process of optimization in Fig.~\ref{fig:t_effectiveness}. From the steps, it can be observed that when $t$
is known to the server, the optimization process is smooth. This demonstrates that generating the changing $t$ as the input of $\mathcal{D}(\hat{\boldsymbol{x}}_0, t)$ influences the approximation during the optimization process of $t$. However, even without knowing $t$, the $128 \times 128$ images can be recovered in the end due to the less randomness than $\boldsymbol{\epsilon}$.

\section{Conclusion}
\label{sec:conclusion}
In this paper, we are the first to study gradient inversion attacks on diffusion models. When the adversarial server knows the Gaussian noise and sampling step for training, we propose \alg to optimize the dummy image in a generative phase followed by a fine-tuning phase. Our experiments demonstrate the importance of both phases for recovering high-resolution data, allowing for reconstruction when other attacks fails.
For the challenging scenario when noise and step are unknown, our proposed \algp further integrates a triple optimization process that coordinates three independent optimizers to conduct optimizations with intervals. \algp is able to invert image size of $32 \times 32$, when all of the baselines fail.

\clearpage
\bibliographystyle{plain}
\bibliography{inversion_diffusion}

\clearpage

\end{document}